\documentclass[letterpaper, 10 pt, conference]{ieeeconf}  

\IEEEoverridecommandlockouts                              

\usepackage{cite}
\usepackage{amsmath,amssymb,amsfonts}
\usepackage{algorithmic}
\usepackage{graphicx}
\usepackage{textcomp}
\usepackage{xcolor}
\usepackage{CJKutf8}
\usepackage{colortbl}
\usepackage{pifont}
\usepackage{empheq}
\usepackage{multirow}
\usepackage{multicol}
\usepackage{booktabs}           
\usepackage{amsfonts, amsmath, amssymb}
\usepackage[ruled, linesnumbered]{algorithm2e}
\usepackage{subfig}
\usepackage{bm}

\usepackage{tabularx}
\newcolumntype{C}[1]{>{\centering\arraybackslash}p{#1}}

\usepackage{fancyhdr}

\fancypagestyle{topline}{
    \fancyhf{}
    \fancyhead[C]{\textbf{Accepted as a conference paper at IEEE SMC 2025}}

    \setlength{\headheight}{14pt}
    \setlength{\headsep}{13pt}      
    \addtolength{\topmargin}{-10pt} 
}
\usepackage{hyperref}
\hypersetup{
    colorlinks=true,
    linkcolor=[rgb]{0.1, 0.15, 0.55},        
    citecolor=[rgb]{0.1, 0.15, 0.55},      
    urlcolor=[rgb]{0.1, 0.15, 0.55}      
}

\title{\LARGE \bf
Relative-Absolute Fusion: Rethinking Feature Extraction in \\Image-Based Iterative Method Selection for \\Solving Sparse Linear Systems
}

\author{Kaiqi Zhang$^{1,2}$, Mingguan Yang$^{2}$, Dali Chang$^{1,2}$, Chun Chen$^{2}$, Yuxiang Zhang$^{2}$, Kexun He$^{1,3}$, Jing Zhao$^{1}$ \!\!\!\!\!\!\!\!\!\!\!\!
	\thanks{$^{1}$\,School of Software Technology, Dalian University of Technology, Dalian, China.
	{\tt\small 
	\{zhangkq, 569377793\}@mail.dlut.edu.cn,
	zhaoj9988@dlut.edu.cn
	}
	}
	\thanks{$^{2}$\,Greater Bay Area National Center of Technology Innovation, Guangzhou, China. 
	{\tt\small 
	\{yangmingguan, chenchun, zhangyuxiang\}@ncti-gba.cn
	}
	}
	\thanks{$^{3}$\,CATARC Automotive Test Center (Tianjin) Co., Ltd, Tianjin, China. 
	{\tt\small 
	hekexun@catarc.ac.cn
	}
	}
}

\begin{document}

\def\mathbi#1{\textbf{\em #1}}

\maketitle

\thispagestyle{topline}

\begin{abstract}

Iterative method selection is crucial for solving sparse linear systems because these methods inherently lack robustness.
Though image-based selection approaches have shown promise, their feature extraction techniques might encode distinct matrices into identical image representations, leading to the same selection and suboptimal method.
In this paper, we introduce RAF (Relative-Absolute Fusion), an efficient feature extraction technique to enhance image-based selection approaches.
By simultaneously extracting and fusing image representations as relative features with corresponding numerical values as absolute features, RAF achieves comprehensive matrix representations that prevent feature ambiguity across distinct matrices, thus improving selection accuracy and unlocking the potential of image-based selection approaches.
We conducted comprehensive evaluations of RAF on SuiteSparse and our developed BMCMat (Balanced Multi-Classification Matrix dataset), demonstrating solution time reductions of 0.08s-0.29s for sparse linear systems, which is 5.86\%-11.50\% faster than conventional image-based selection approaches and achieves state-of-the-art (SOTA) performance.
BMCMat is available at \url{https://github.com/zkqq/BMCMat}.  

\end{abstract}

\section{Introduction}

Solving sparse linear systems, as shown in Eq. \ref{Ax=b}, is a fundamental task in scientific computing.
\begin{equation}
	\label{Ax=b}
	\mathcal{A}x=b
\end{equation}
where $\mathcal{A} \in \mathbb{R}^{n \times n}$ denotes a sparse coefficient matrix, $b \in \mathbb{R}^{n}$ represents the right-hand side vector, and $x \in \mathbb{R}^{n}$ is the unknown solution vector \cite{zou2023survey}.
Such systems are typically solved through either direct methods \cite{davis2016survey} or iterative methods \cite{saad2003iterative}.
For large-scale systems, solving Eq. \ref{Ax=b} is computationally intensive, especially accounting for over 70\% of the total computation time in reservoir engineering \cite{gasparini2021hybrid}.
This computational burden has prompted the widespread adoption of iterative methods, which inherently exhibit lower computational complexity.
However, iterative methods demonstrate limited robustness.
An appropriate iterative method can solve the system efficiently, whereas an unsuitable one may result in slow convergence or divergence.
Unfortunately, selecting an optimal (fastest convergence) iterative method for a given linear system remains challenging, often relying on trial and error or expert intuition \cite{scott2023algorithms}.

\begin{figure}[t]
	\vspace{1.7mm}
    \centering
    \includegraphics[width=0.485\textwidth]{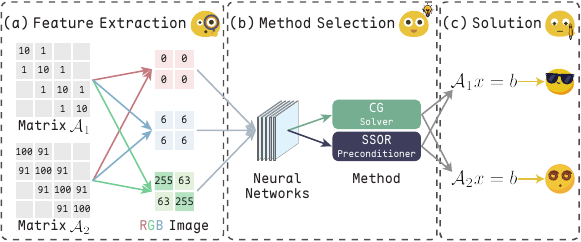}
    \vspace{-5mm}
    \caption{
    Motivation for RAF.
    Conventional feature extraction techniques in image-based selection approaches encode distinct matrices $\mathcal{A}_1$ and $\mathcal{A}_2$ into identical RGB image representations, yielding the same selection method (\textit{CG \& SSOR}).
    Although the \textit{CG \& SSOR} method efficiently solves $\mathcal{A}_1x=b$, it exhibits poor performance for $\mathcal{A}_2x=b$.
    }
    \label{motivation}
    \vspace{-3.7mm}
\end{figure}

Deep learning advances have driven research into selecting optimal iterative methods for sparse linear systems based on the features of matrix $\mathcal{A}$.
Early studies utilized Fully Connected (FC) networks to select optimal methods based on manually identified numerical features from $\mathcal{A}$ \cite{funk2022prediction}.
Subsequent approaches modeled $\mathcal{A}$ as a topological graph and employed Graph Neural Network (GNN) for selection \cite{tang2022graph}.
The SOTA work involves image-based selection, which encodes $\mathcal{A}$ as an RGB image and applies Convolutional Neural Network (CNN) for efficient selection \cite{yamada2018preconditioner, souza2023comparison}.

Although image-based selection approaches achieve SOTA performance, their feature extraction techniques may encode distinct matrices into identical RGB image representations, potentially leading to suboptimal method selection.
As illustrated in Fig. \ref{motivation}, although \textit{CG \& SSOR} is deemed optimal for both matrices $\mathcal{A}_1$ and $\mathcal{A}_2$, it is only truly appropriate for the strongly diagonally dominant matrix $\mathcal{A}_1$, whereas the weakly diagonally dominant matrix $\mathcal{A}_2$ is solved more efficiently using \textit{CG \& $\omega$-Jacobi}.
This limitation stems from feature extraction techniques that compute RGB image representations, incorporating only relative matrix features while omitting absolute ones.
For instance, the red channel computation biases non-zero elements by subtracting the matrix's minimum value (Eq. \ref{eq: v}) and subsequently normalizes the results (Eq. \ref{eq: red}), thus capturing only the relative magnitude relationships between matrix elements.
The exclusion of absolute features may allow matrices with varying magnitudes to yield identical red channels, leading to feature ambiguity and limiting the effectiveness of image-based selection approaches.

Rethinking feature extraction techniques in image-based selection approaches, we propose RAF, an efficient feature extraction technique that addresses existing limitations.
RAF extracts both image representations as relative features and corresponding numerical values as absolute features, subsequently fusing these complementary features to achieve complete matrix representations and eliminate feature ambiguity across distinct matrices.
For instance, when computing the red channel, RAF simultaneously extracts the matrix's minimum value as a bias reference, which characterizes matrix magnitude as an absolute feature, ensuring comprehensive representation of non-zero element magnitudes after fusion, preventing matrices with varying magnitudes from yielding identical features.
Our proposed RAF mitigates information loss typically associated with purely relative features and enhances the efficiency of image-based selection approaches.

Our contributions can be summarized as follows:

\begin{itemize}
	\item We developed BMCMat based on Partial Differential Equation (PDE) discretization to mitigate label imbalance in the widely used SuiteSparse dataset \cite{davis2011university}, facilitating research on iterative method selection for sparse linear systems ($\S$ \ref{sec: dataset}).
	
	\item We introduce RAF, an efficient feature extraction technique that extracts and fuses relative and absolute matrix features to eliminate feature ambiguity across matrices, unlocking the potential of image-based method selection. ($\S$ \ref{sec: raf}, \ref{sec: pipeline})
	
	\item We comprehensively evaluated RAF on SuiteSparse and BMCMat, demonstrating selection accuracy improvements of 0.022-0.039 and solution time reductions of 0.08s-0.29s for linear systems, which is 5.86\%-11.50\% faster than conventional image-based selection approaches. To our best knowledge, RAF achieves SOTA performance in method selection ($\S$ \ref{sec: eff}).
\end{itemize}

\section{Methodology}

The efficacy of deep learning fundamentally depends on two critical factors: high-quality training data and effective algorithm design.
This section enhances iterative method selection through dual innovations: improved dataset and optimized model.

\subsection{Problem Formulation}

Iterative method selection fundamentally establishes a mapping $f$ from a given matrix $\mathcal{A}$ to the optimal method, as shown in Eq. \ref{y=f(a)}.
\begin{equation}
	\label{y=f(a)}
	y = f(\mathcal{A})
\end{equation}
where $y$ denotes an iterative method, comprising a solver and a preconditioner.
Consequently, method selection can be formulated as a multi-class classification problem.
For a set of linear systems with $k$ available iterative methods, each matrix $\mathcal{A}$ is assigned an $k$-dimensional label vector $\{0, 1\}^k$, where only the element corresponding to the optimal method is $1$, and the remaining $k-1$ entries, corresponding to suboptimal methods, are $0$.

\begin{table}[t]
	\vspace{1.2mm}
	\centering
	\caption{Common iterative methods comprising various solvers and preconditioners \cite{petsc_solvers_doc, zou2023survey}.}
	\vspace{-2mm}
	\label{tab: iter}

	\begin{tabular}{*2{c}|*2{c}}
	
	\toprule
		
	\rowcolor{gray!8} \multicolumn{4}{c}{\textbf{Iterative Methods}} \\

	\rowcolor{gray!8} \multicolumn{2}{c}{Solvers} & \multicolumn{2}{c}{Preconditioners} \\

	\midrule

	CG & F-CG & $\omega$-Jacobi & Blocked Jacobi \\
	
	GMRES & F-GMRES & G-S & SSOR \\
	
	L-GMRES & BICG & GMG & AMG \\
	
	GCR & BICGSTAB & DDM & ILU \\
	
	\bottomrule
	
	\end{tabular}
	\vspace{-5mm}
\end{table}

\begin{figure}[!t]
	\centering
	\subfloat[SuiteSparse.]
	{
		\includegraphics[width=0.44\linewidth]{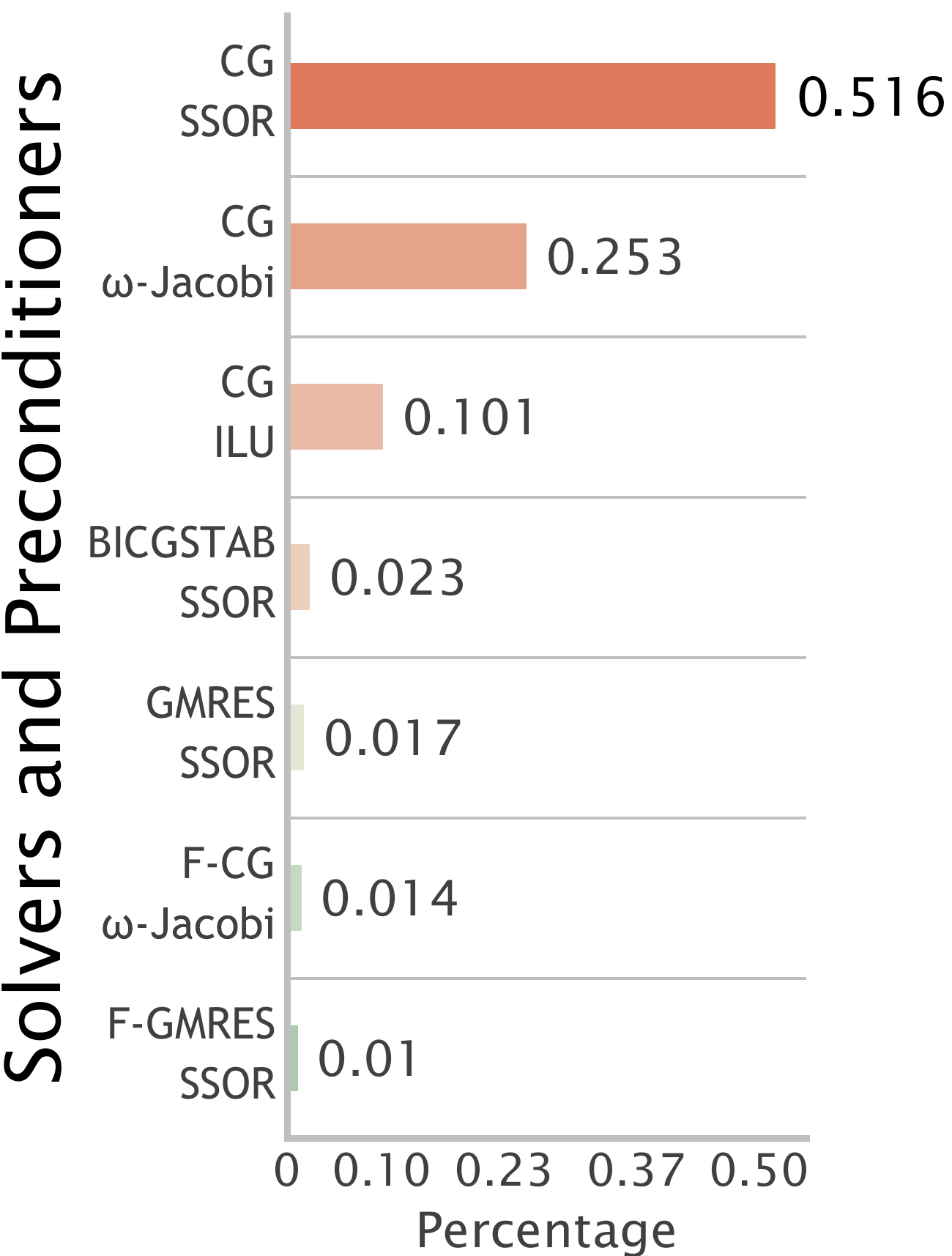}
		\label{fig: ss}
	} 
 	\hspace{-0.85cm}
	\subfloat[BMCMat.]
	{
		\includegraphics[width=0.44\linewidth]{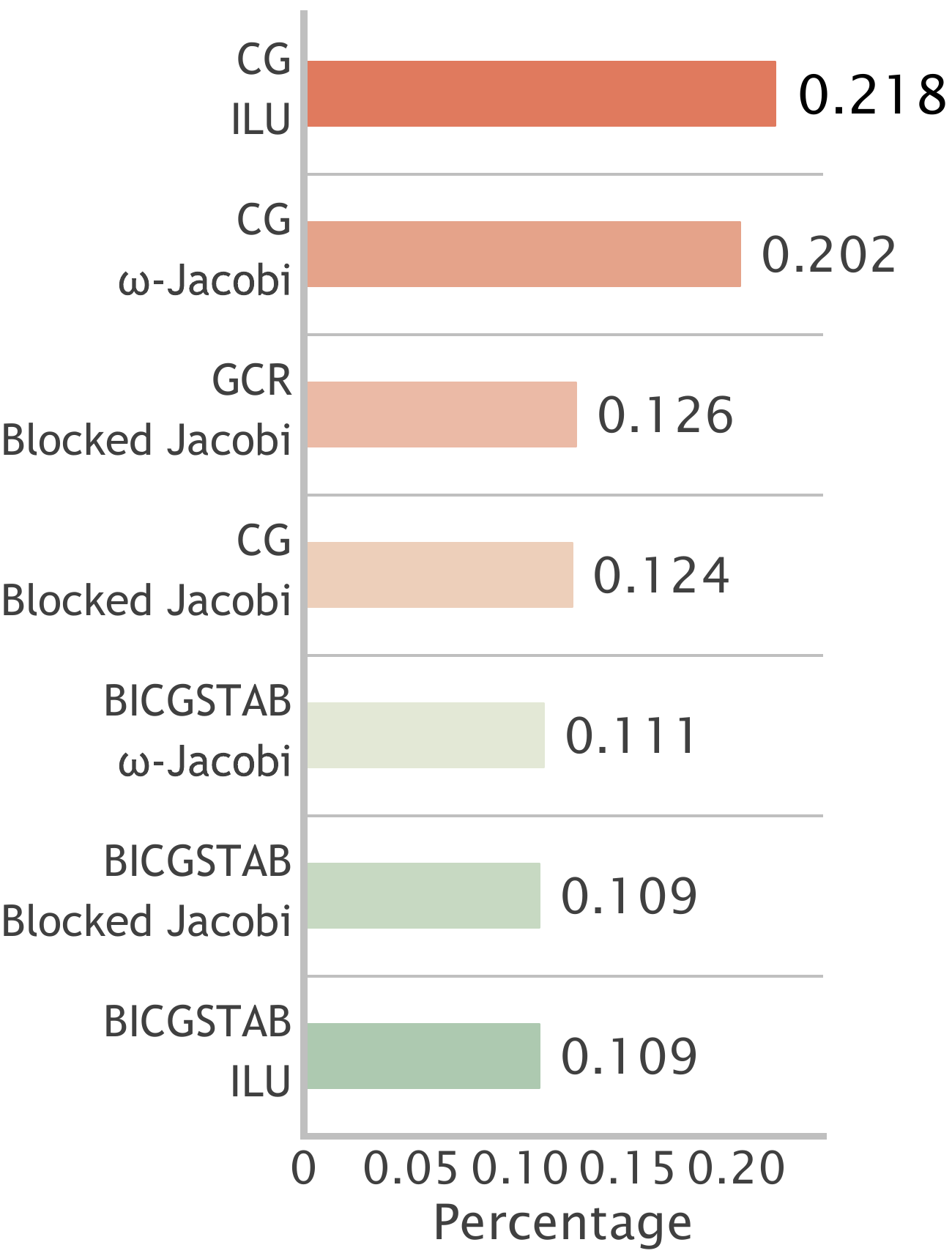}
		\label{fig: bmc}
	}
	\vspace{-1mm}
	\caption{
	Distribution of optimal iterative methods across datasets.
	For clarity, only seven methods with the highest percentages are listed for the SuiteSparse dataset.
	}
	\label{}
	\vspace{-2mm}
\end{figure}

\subsection{BMCMat}
\label{sec: dataset}

Effective $k$-class classification in deep learning relies on high-quality training data due to its inherently data-driven nature.

We initially examined the widely adopted SuiteSparse dataset, from which we extracted 576 matrices $\mathcal{A} \in \mathbb{R}^{n \times n}$ with $1000 \leq n \leq 10000$ \cite{tang2022graph}. 
Using common iterative methods from the PETSc library \cite{petsc_solvers_doc, zou2023survey} listed in Table \ref{tab: iter}, we determined the optimal method for each matrix.
Notably, although 64 ($8 \times 8$) theoretical solver-preconditioner combinations are possible, only a subset was computationally feasible, and an even smaller fraction demonstrated optimal performance.
Finally, the method selection was formulated as a 25-class classification task, as illustrated in Fig. \ref{fig: ss}, revealing significant class imbalance within the SuiteSparse. 
The \textit{CG \& SSOR} method constitutes over 50\% of all cases, while 19 other methods individually represent less than 1\% of the data.
This imbalance introduces a selection bias toward dominant classes (e.g., \textit{CG \& SSOR}). 
Although such bias may yield high selection accuracy, it significantly compromises performance on minority classes, leading to suboptimal method selection for specific linear systems.

To address this limitation, we developed BMCMat, a relatively balanced multi-class classification Matrix dataset.
Specifically, we generated 50,000 linear systems $\mathcal{A}x=b$ with $1000 \leq n \leq 10000$ using OpenMat \cite{OpenMat2024}, a parallel sparse matrix generator based on PDE discretization, and applied iterative methods in Table \ref{tab: iter} to identify the optimal method for each system.
We then extracted 3,819 matrices from these systems to construct BMCMat, which contains seven distinct method classes, as illustrated in Fig. \ref{fig: bmc}.
Compared to SuiteSparse, BMCMat exhibits a more balanced class distribution, with a maximum class ratio of only 2:1.
This balanced distribution improves feature learning across method classes, thereby enhancing the model's ability to accurately select optimal methods for diverse linear systems in practical applications.

\subsection{RAF}
\label{sec: raf}

In addition to high-quality training data, an efficient algorithm is crucial for $k$-class classification in deep learning.

Conventional feature extraction techniques in image-based selection approaches involve three sequential steps: (1) defining image resolution $m$, (2) partitioning matrix $\mathcal{A}$ into $m^2$ blocks, and (3) computing RGB channels for each pixel, representing its corresponding block.
Each RGB channel captures distinct matrix features: the red channel represents non-zero element magnitudes (Eqs. \ref{eq: red}, \ref{eq: r}, \ref{eq: v}, \ref{eq: d}), the blue channel encodes matrix dimensions (Eq. \ref{eq: blue}), and the green channel quantifies non-zero element density (Eq. \ref{eq: green}) \cite{yamada2018preconditioner, souza2023comparison}.
\begin{subequations}
\begin{empheq}[left=\empheqlbrace]{align}
	R_{ij} &= \left\lfloor \frac{\gamma_{ij} - \min(\gamma)}{\max(\gamma) - \min(\gamma)} \times 255 \right\rfloor \label{eq: red} \\
    &\gamma_{ij} = \left\{ \begin{aligned}
&\frac{\sum_{a \in A_{ij}} v(a)}{NNZ_{ij}}, & \delta \leq 255 \\
&\frac{\sum_{a \in A_{ij}} \log_2 v(a)}{NNZ_{ij}}, & \delta > 255
\end{aligned} \right. \label{eq: r} \\
& v(a) = a - \min(\mathcal{A}) + 1 \label{eq: v} \\
& \delta = \max(\mathcal{A}) - \min(\mathcal{A}) \label{eq: d} \\
    B_{ij} &= \left\lfloor \frac{N_{\mathcal{A}} - N_{\min}}{N_{\max} - N_{\min}} \times 255 \right\rfloor \label{eq: blue} \\
    G_{ij} &= \left\lfloor \frac{NNZ_{ij}}{{N_b}^2} \times 255 \right\rfloor \label{eq: green} 
\end{empheq}
\label{rgb}
\end{subequations}

Here, indices $i, j \in \{1,\ldots,m\}$ identify specific blocks within the partitioned matrix.
$\delta$ represents the matrix value range, with $\min(\mathcal{A})$ and $\max(\mathcal{A})$ denoting the minimum and maximum values, respectively. 
$v(a)$ denotes the biased value of matrix element $a$.
$NNZ_{ij}$ counts the number of non-zero elements in block $\mathcal{A}_{ij}$, while $\gamma_{ij}$ represents their biased average.
$N_{\mathcal{A}}$ indicates the matrix order, while $N_{\min}$ and $N_{\max}$ denote the minimum and maximum matrix orders in the dataset, respectively.
The order of each block, $N_b$, is approximated by $N_b \approx N_{\mathcal{A}} / m$.

Such feature extraction techniques that focus solely on relative relationships cannot completely represent matrices, leading to identical relative feature extraction for distinct matrices due to the absence of critical absolute features (Fig. \ref{motivation}), thus limiting the effectiveness of image-based selection approaches.
This limitation arises primarily because matrices with varying magnitudes but identical ranges may yield identical red channel values owing to the bias (Eq. \ref{eq: v}) and normalization (Eq. \ref{eq: red}).
Furthermore, linear and logarithmic block-wise averaging can generate identical $\gamma$ values despite different biased inputs $v(a)$, further contributing to feature ambiguity (Eqs. \ref{eq: r}, \ref{eq: d}).
Moreover, the reliance of the blue and green channels on the values $N_{\min}$, $N_{\max}$, and $N_b$ limits the accuracy of matrix representation.

\begin{figure}[t]
	\vspace{1.4mm}
    \centering
    \includegraphics[width=0.48\textwidth]{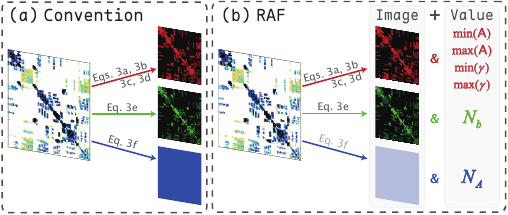}
    \vspace{-4.5mm}
    \caption{
Comparison between conventional feature extraction techniques (left) and RAF (right) in image-based selection approaches.
RAF simultaneously extracts relative image representations and corresponding absolute numerical values, yielding comprehensive matrix features for enhanced characterization.
Notably, because $N_{\mathcal{A}}$ demonstrates higher effectiveness than the blue channel, RAF retains only the red and green channels in the image representation.
    }
    \label{raf}
    \vspace{-2.6mm}
\end{figure}

\begin{figure*}[t]
    \centering
    \includegraphics[width=0.88\textwidth]{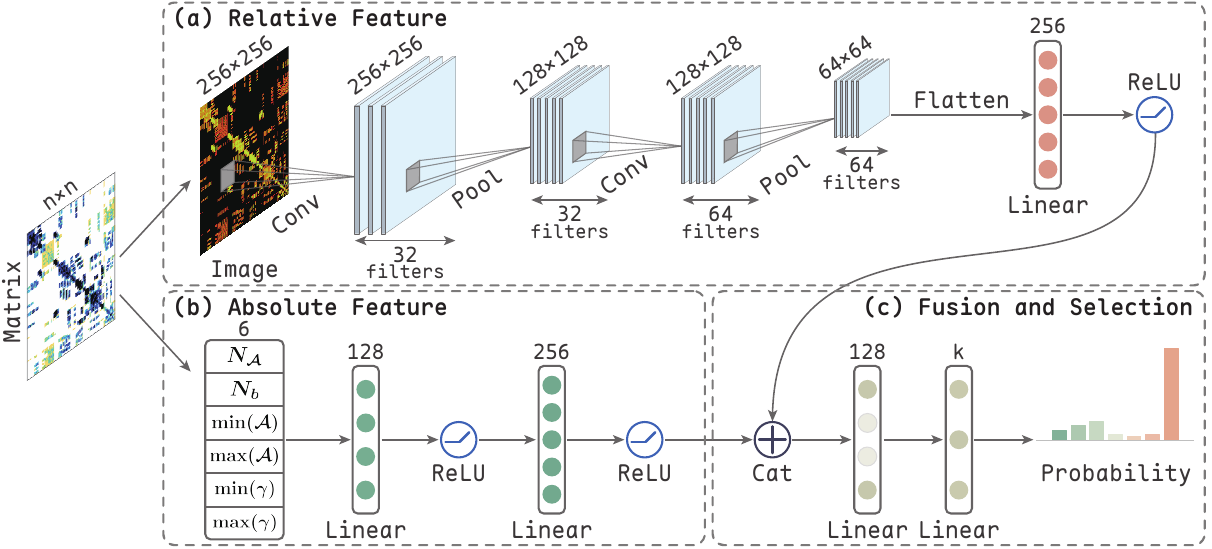}
    \vspace*{-1mm}
    \caption{
    Pipeline for image-based iterative method selection with RAF.
    Component (a) extracts the red and green channels as relative matrix features, which are subsequently learned via convolution to capture spatial patterns within the matrix.
    Component (b) simultaneously extracts six corresponding numerical values as absolute matrix features and learns them through linear layers.
    Finally, component (c) fuses these features to select the optimal iterative method for the given matrix.
    }
    \label{pipeline}
    \vspace*{-4mm}
\end{figure*}

To address these limitations, we propose RAF, an efficient feature extraction technique for enhancing image-based selection approaches.
As illustrated in Fig. \ref{raf}, RAF employs a dual-feature strategy that simultaneously extracts image representations as relative features along with corresponding numerical values as absolute features to create comprehensive matrix representations.
For non-zero element magnitudes, RAF extracts both the red channel and the absolute boundary values $\min(\mathcal{A})$, $\max(\mathcal{A})$, $\min(\gamma)$, and $\max(\gamma)$ (Eqs. \ref{eq: red}, \ref{eq: r}) for complete characterization.
For dimensionality, RAF uses the matrix order $N_{\mathcal{A}}$ (Eq. \ref{eq: blue}) to fully characterize the matrix size.
Interestingly, since $N_{\mathcal{A}}$ provides complete dimensional information, RAF eliminates the blue channel to reduce feature redundancy.
For non-zero element density, RAF extracts both the green channel and the block order $N_b$ (Eq. \ref{eq: green}) to characterize the sparsity pattern.
For the matrices $\mathcal{A}_1$ and $\mathcal{A}_2$ shown in Fig. \ref{motivation}, when extracting the red and green channels, RAF also extracts $\min(\mathcal{A}_1)=1$, $\max(\mathcal{A}_1)=10$, $\min(\mathcal{A}_2)=91$, and $\max(\mathcal{A}_2)=100$ to distinguish between them and avoid feature ambiguity. 
Additionally, RAF extracts $\min(\gamma)=1$, $\max(\gamma)=5.5$, $N_b=2$, and $N_{\mathcal{A}}=4$ for a more precise characterization of the matrices.
Finally, relative and absolute features are fused to select the optimal iterative method for linear systems ($\S$ \ref{sec: pipeline}).
By extracting and fusing relative image representations with absolute numerical values, RAF achieves more complete matrix representations, enhancing image-based selection approaches to better differentiate matrices and improve selection accuracy.

\subsection{Image-Based Method Selection with RAF}
\label{sec: pipeline}

After introducing the efficient RAF, the next step is to consider its integration into the image-based method selection architecture.
 
As illustrated in Fig. \ref{pipeline}, the image-based method selection with RAF comprises three components: (a) extracting and learning the matrix's relative features, (b) extracting and learning the matrix's absolute features, and (c) fusing features to select the optimal iterative method for the input matrix.
As RAF, component (a) exclusively extracts red and green channel image representations as the matrix's relative features. 
The red and green channels of the $256 \times 256$ pixel image undergo transformed into a $64 \times 64 \times 64$ tensor via sequential operations: a $3 \times 3$ convolution (Conv) with 32 filters, $2 \times 2$ max pooling (Pool), followed by a $3 \times 3$ convolution with 64 filters, and a final $2 \times 2$ max pooling.
To match the dimension of the absolute feature vector, this tensor is subsequently flattened and reduced to a 256-dimensional vector through a linear transformation with ReLU activation.
Simultaneously, component (b) extracts six corresponding numerical values representing the matrix's absolute features, expanding this feature vector to 256 dimensions via two sequential linear transformation layers, each followed by ReLU activation.
Finally, in component (c), the two 256-dimensional vectors (relative and absolute features) are concatenated (Cat) to form a unified 512-dimensional feature vector.
This combined vector passes through two linear layers, with a dropout rate of 0.5 applied to the first layer to mitigate overfitting. 
The second layer contains $k$ neurons, where $k$ corresponds to the number of candidate methods in the dataset (SuiteSparse: $k=25$; BMCMat: $k=7$).
The output represents probability distributions across all candidate methods, wherein the method exhibiting the highest probability is selected as optimal for the given matrix.

\vspace{1mm}

\section{Experiment}

\subsection{Experimental Setup}

\textbf{Models and parameters.}
We evaluated four iterative method selection models: existing FC \cite{funk2022prediction}, GNN \cite{tang2022graph}, CNN \cite{yamada2018preconditioner, souza2023comparison}, and our proposed image-based approach with RAF (simplified as RAF).
All models were trained on an A6000 GPU using SuiteSparse \cite{davis2011university} and BMCMat.
For RAF, we employed a learning rate of 8e-4, a batch size of 64, and early stopping with a maximum of 100 epochs.
All linear systems were solved using the PETSc library on an Intel Xeon 8352V CPU with 256 GB RAM.

\textbf{Metrics.}
We evaluated model performance using the following quantitative metrics:
\begin{itemize}
	\item Solution time (s): The walltime for solving linear systems using model-selected iterative methods.
	 	
	\item Slowdown: The ratio of optimal method computation time to the model-selected method time. Values closer to 1 indicate the selected method optimal performance.
	
	\item Selection Accuracy: The probability that the model correctly selects the optimal method.
	
	\item Top-$n$ Selection Accuracy: The probability that the optimal method is included in the model's top $n$ selections. Higher values indicate the model's ability to rank the optimal method among its top choices, demonstrating practical effectiveness.
\end{itemize}


\subsection{Effectiveness}
\label{sec: eff}

\begin{table}[!t]
	\vspace{1.3mm}
	\centering
	\caption{
		Solution time and slowdown of existing models and \textbf{RAF} on SuiteSparse and BMCMat.	
	}
	\vspace{-1mm}
	\label{eff: time and slowdown}
	
	\begin{tabular}{*1{c}*1{c}{c}*2{c}}
	
	\toprule
	
	\rowcolor{gray!8} \multicolumn{1}{c}{} & \multicolumn{2}{c}{\textbf{SuiteSparse}} & \multicolumn{2}{c}{\textbf{BMCMat}} \\
	\rowcolor{gray!8} \multicolumn{1}{c}{\textbf{Model}} & Time\textcolor[rgb]{0.847, 0.192, 0.082}{$\, \boldsymbol{\downarrow} \!$} & \multicolumn{1}{c}{Slowdown\textcolor[rgb]{0.002, 0.465, 0.71}{$\, \boldsymbol{\uparrow} \!\!\!\!\!\!\!$}} & Time\textcolor[rgb]{0.847, 0.192, 0.082}{$\, \boldsymbol{\downarrow} \!$} & Slowdown\textcolor[rgb]{0.002, 0.465, 0.71}{$\, \boldsymbol{\uparrow} \!\!\!\!\!\!\!$} \\
	
	\midrule
	
	FC & 0.70	& 0.31	& 5.18	& 0.77 \\
	
	GNN & 0.70 & 0.31 & 5.45 & 0.73 \\
	
	CNN & 0.68\textcolor[rgb]{0.4, 0.4, 0.4}{\scriptsize{ (Base)}}  & 0.32\textcolor[rgb]{0.4, 0.4, 0.4}{\scriptsize{ (Base)}}	& 4.95\textcolor[rgb]{0.4, 0.4, 0.4}{\scriptsize{ (Base)}}	& 0.81\textcolor[rgb]{0.4, 0.4, 0.4}{\scriptsize{ (Base)}} \\
	
	\midrule

	\textbf{RAF} & \textbf{0.61}\textcolor[rgb]{0.847, 0.192, 0.082}{\scriptsize{ (-0.07)}} & \textbf{0.36}\textcolor[rgb]{0.002, 0.465, 0.71}{\scriptsize{ (+0.04)}} & \textbf{4.66}\textcolor[rgb]{0.847, 0.192, 0.082}{\scriptsize{ (-0.29)}} & \textbf{0.86}\textcolor[rgb]{0.002, 0.465, 0.71}{\scriptsize{ (+0.05)}} \\
	
	\bottomrule

	\end{tabular}
	\vspace{-2mm}
\end{table}

\begin{figure}[!t]
	\centering
	\includegraphics[width=0.41\textwidth]{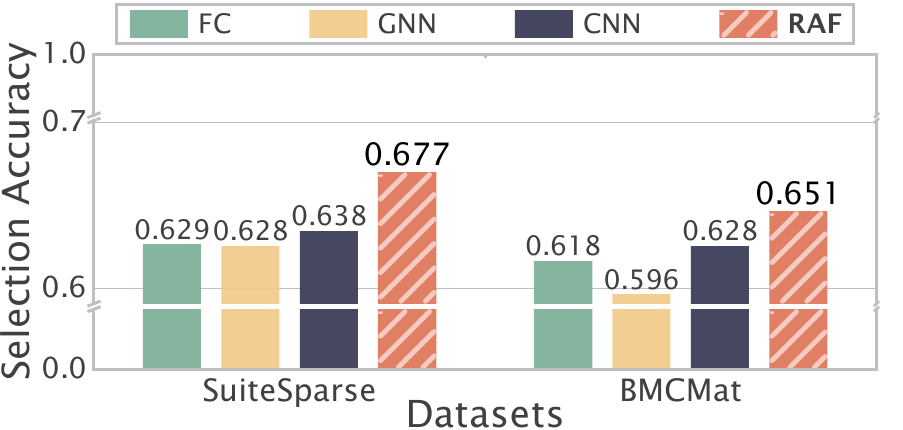}
	\vspace{-1mm}
	\caption{
		Selection accuracy of FC, CNN, GNN, and \textbf{RAF} on SuiteSparse and BMCMat.
		}
	\label{eff: acc}
	\vspace{-3mm}
\end{figure}

\begin{figure}[!t]
	\centering
	\includegraphics[width=0.41\textwidth]{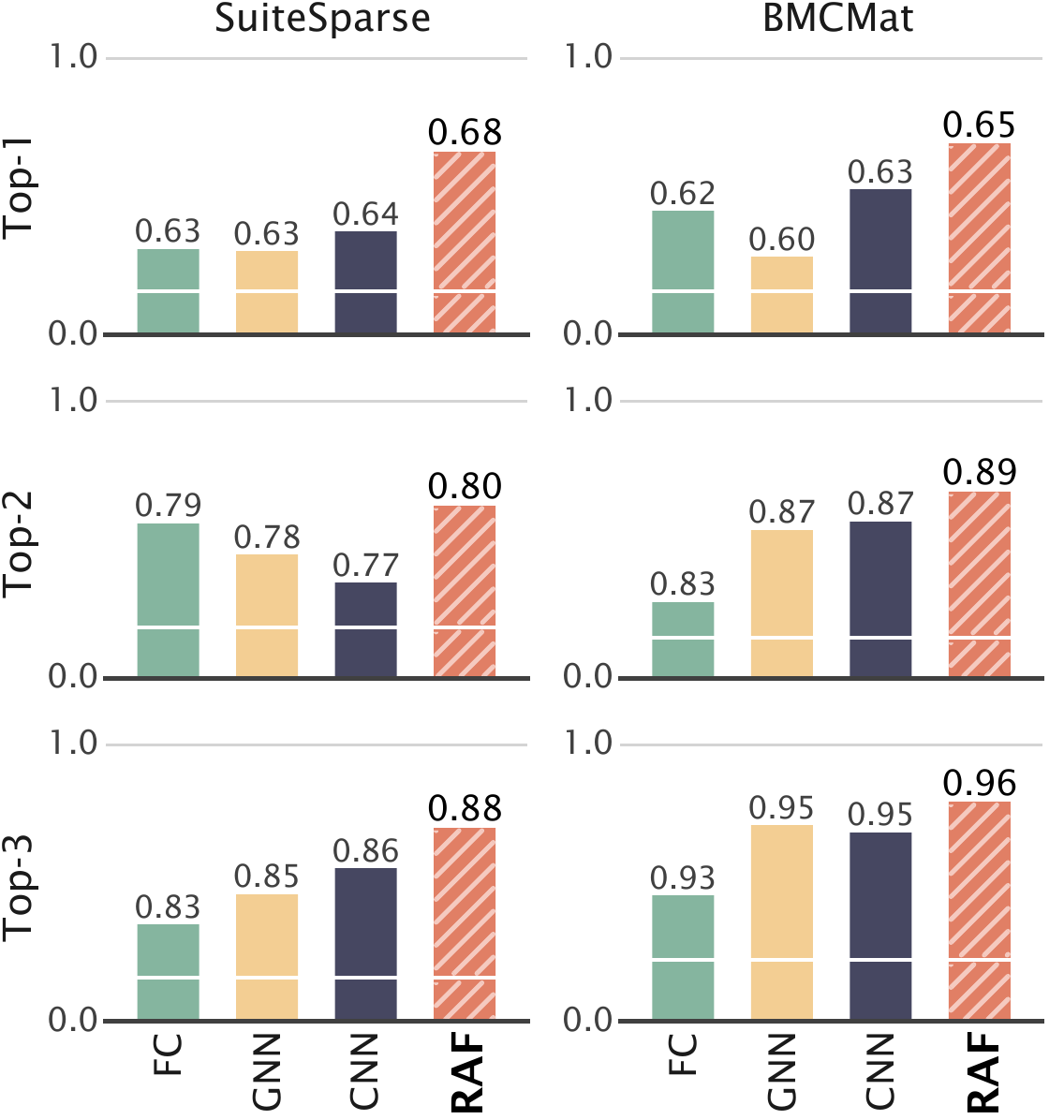}
	\vspace{-1mm}
	\caption{
	Top-$n$ selection accuracy of FC, CNN, GNN, and \textbf{RAF} on SuiteSparse and BMCMat.
	}
	\label{eff: top-acc}
	\vspace{-5mm}
\end{figure}

Table \ref{eff: time and slowdown} compares the solution times of four iterative method selection approaches on the SuiteSparse and BMCMat datasets.
RAF demonstrates SOTA performance, achieving the fastest solution times among all evaluated approaches.
On SuiteSparse, RAF reduces solution times by 0.08s-0.10s compared to competing models, yielding speedups of 1.13x-1.16x.
For BMCMat, RAF decreases solution times by 0.29s-0.79s relative to alternative approaches, achieving speedups of 1.06x-1.17x.
RAF's superior performance can be attributed to its ability to more effectively extract matrix features compared to conventional image-based approaches, thus charactering matrix completely and avoiding feature ambiguity.
Table \ref{eff: time and slowdown} further validates these findings by comparing the slowdown across all four approaches on both datasets.
RAF achieves optimal slowdown of 0.36 and 0.86 on SuiteSparse and BMCMat, respectively, outperforming competitors by margins of 0.04-0.12, confirming its selection of near-optimal methods and further demonstrating its superiority.

\begin{figure}[!t]
	\centering
	\includegraphics[width=0.45\textwidth]{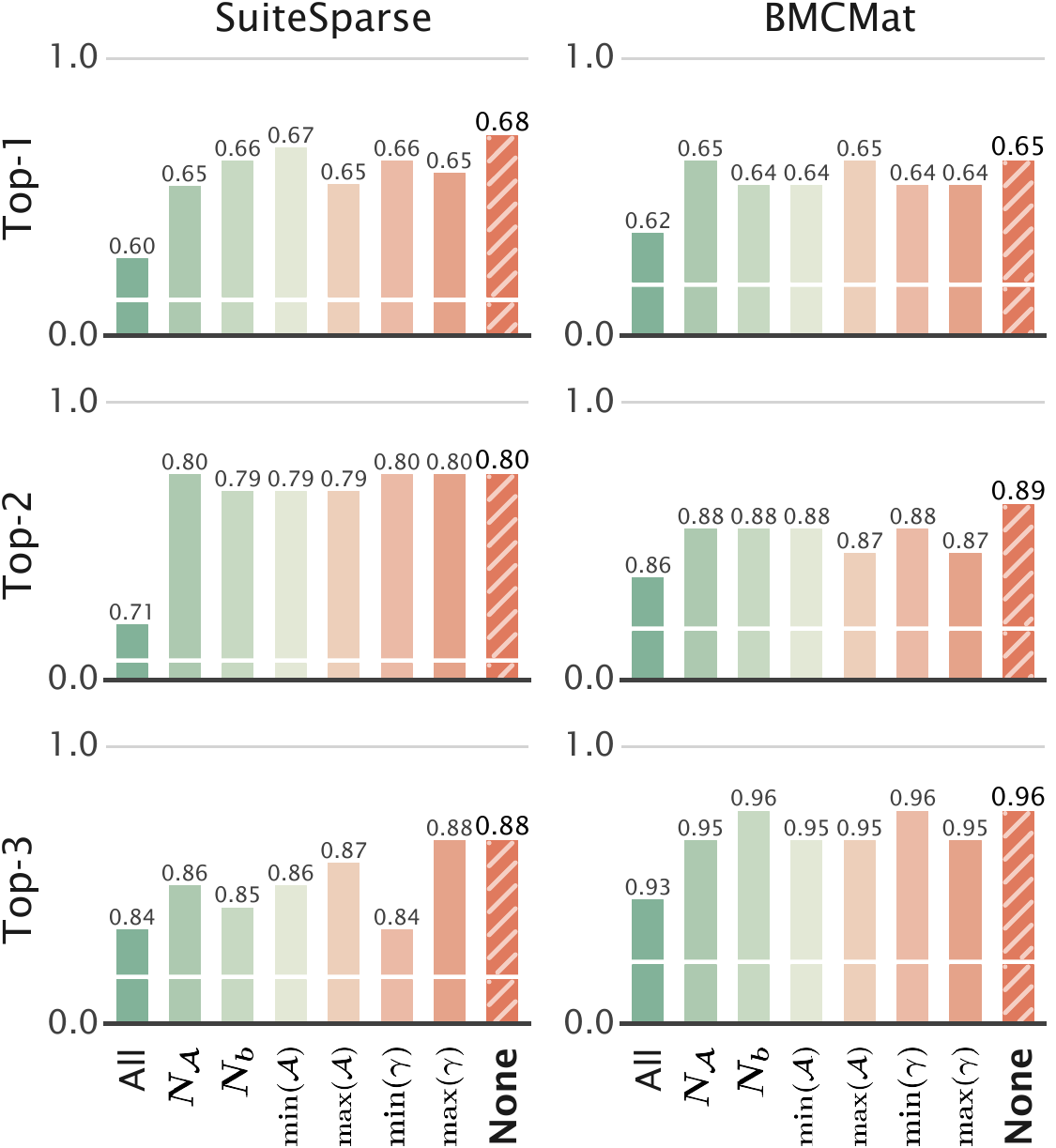}
	\vspace{-1.5mm}
	\caption{
	Top-$n$ selection accuracy of various \textbf{RAF} variants on SuiteSparse and BMCMat.
	The horizontal axis represents different configurations of numerical value exclusions.
	``All" refers to the \textbf{RAF} variant with all numerical values removed and ``\textbf{None}" represents the complete \textbf{RAF}.
	}
	\label{ab: fig}
	\vspace{-5mm}
\end{figure}

\begin{table*}[!t]
	\centering
	\caption{
Ablation study results showing selection accuracy, speedup, and slowdown for \textbf{RAF} variants on SuiteSparse and BMCMat.
Numerical values are represented by symbols, with ``\textcolor[rgb]{0.002, 0.465, 0.71}{\ding{51}}'' signifying presence and ``\textcolor[rgb]{0.847, 0.192, 0.082}{\ding{55}}'' signifying absence.
The \colorbox[rgb]{0.885, 0.96, 1.0}{best} and \colorbox{red!7}{worst} performance are illustrated with corresponding colors.
Results of complete \textbf{RAF} are displayed in \textbf{bold}.
}

	\vspace*{-1mm}
	
	\label{ab: table}
	
	\begin{tabular}{*1{C{1.5cm}|}*6{C{0.95cm}}|c*2{C{2cm}}}
	
	\toprule
	
	\rowcolor{gray!8} \multicolumn{1}{c}{} & \multicolumn{6}{c}{\textbf{Numerical Value}} & \multicolumn{3}{c}{\textbf{Metric}} \\
	
	\rowcolor{gray!8} \multicolumn{1}{c}{\textbf{Dataset}} & ${N_\mathcal{A}}$ & ${N_b}$ &  ${\min(\mathcal{A})}$ &  ${\max(\mathcal{A})}$  &  ${\min(\gamma)}$  & \multicolumn{1}{c}{${\max(\gamma)}$}  &{Selection accuracy\textcolor[rgb]{0.002, 0.465, 0.71}{$\, \boldsymbol{\uparrow} \!\!\!\!\!$}} & {Solution time}\textcolor[rgb]{0.847, 0.192, 0.082}{$\, \boldsymbol{\downarrow} \!$} &  {Slowdown}\textcolor[rgb]{0.002, 0.465, 0.71}{$\, \boldsymbol{\uparrow} \!\!\!\!\!\!\!$} \\
	
	\midrule
	
	\multirow{8}{*}{\textbf{SuiteSparse}} & \textcolor[rgb]{0.002, 0.465, 0.71}{\ding{51}} & \textcolor[rgb]{0.002, 0.465, 0.71}{\ding{51}} & \textcolor[rgb]{0.002, 0.465, 0.71}{\ding{51}} & \textcolor[rgb]{0.002, 0.465, 0.71}{\ding{51}} & \textcolor[rgb]{0.002, 0.465, 0.71}{\ding{51}} & \textcolor[rgb]{0.002, 0.465, 0.71}{\ding{51}} & \cellcolor[rgb]{0.885, 0.96, 1.0}\textbf{0.677}\textcolor[rgb]{0.4, 0.4, 0.4}{\scriptsize{ (Base)}} & \cellcolor[rgb]{0.885, 0.96, 1.0}\textbf{0.605}\textcolor[rgb]{0.4, 0.4, 0.4}{\scriptsize{ (Base)}} & \cellcolor[rgb]{0.885, 0.96, 1.0}\textbf{0.364}\textcolor[rgb]{0.4, 0.4, 0.4}{\scriptsize{ (Base)}} \\
	
	& \textcolor[rgb]{0.847, 0.192, 0.082}{\ding{55}} & \textcolor[rgb]{0.002, 0.465, 0.71}{\ding{51}} & \textcolor[rgb]{0.002, 0.465, 0.71}{\ding{51}} & \textcolor[rgb]{0.002, 0.465, 0.71}{\ding{51}} & \textcolor[rgb]{0.002, 0.465, 0.71}{\ding{51}} & \textcolor[rgb]{0.002, 0.465, 0.71}{\ding{51}} & 0.647\textcolor[rgb]{0.847, 0.192, 0.082}{\scriptsize{ (-0.030)}} & 0.651\textcolor[rgb]{0.002, 0.465, 0.71}{\scriptsize{ (+0.046)}} & 0.338\textcolor[rgb]{0.847, 0.192, 0.082}{\scriptsize{ (-0.026)}} \\
	
	& \textcolor[rgb]{0.002, 0.465, 0.71}{\ding{51}} & \textcolor[rgb]{0.847, 0.192, 0.082}{\ding{55}} & \textcolor[rgb]{0.002, 0.465, 0.71}{\ding{51}} & \textcolor[rgb]{0.002, 0.465, 0.71}{\ding{51}} & \textcolor[rgb]{0.002, 0.465, 0.71}{\ding{51}} & \textcolor[rgb]{0.002, 0.465, 0.71}{\ding{51}} & 0.661\textcolor[rgb]{0.847, 0.192, 0.082}{\scriptsize{ (-0.016)}} & 0.622\textcolor[rgb]{0.002, 0.465, 0.71}{\scriptsize{ (+0.017)}} & 0.354\textcolor[rgb]{0.847, 0.192, 0.082}{\scriptsize{ (-0.010)}} \\
	
	& \textcolor[rgb]{0.002, 0.465, 0.71}{\ding{51}} & \textcolor[rgb]{0.002, 0.465, 0.71}{\ding{51}} & \textcolor[rgb]{0.847, 0.192, 0.082}{\ding{55}} & \textcolor[rgb]{0.002, 0.465, 0.71}{\ding{51}} & \textcolor[rgb]{0.002, 0.465, 0.71}{\ding{51}} & \textcolor[rgb]{0.002, 0.465, 0.71}{\ding{51}} & 0.669\textcolor[rgb]{0.847, 0.192, 0.082}{\scriptsize{ (-0.008)}} & 0.611\textcolor[rgb]{0.002, 0.465, 0.71}{\scriptsize{ (+0.006)}} & 0.360\textcolor[rgb]{0.847, 0.192, 0.082}{\scriptsize{ (-0.004)}} \\
	
	& \textcolor[rgb]{0.002, 0.465, 0.71}{\ding{51}} & \textcolor[rgb]{0.002, 0.465, 0.71}{\ding{51}} & \textcolor[rgb]{0.002, 0.465, 0.71}{\ding{51}} & \textcolor[rgb]{0.847, 0.192, 0.082}{\ding{55}} & \textcolor[rgb]{0.002, 0.465, 0.71}{\ding{51}} & \textcolor[rgb]{0.002, 0.465, 0.71}{\ding{51}} & 0.648\textcolor[rgb]{0.847, 0.192, 0.082}{\scriptsize{ (-0.029)}} &	0.645\textcolor[rgb]{0.002, 0.465, 0.71}{\scriptsize{ (+0.040)}} & 0.341\textcolor[rgb]{0.847, 0.192, 0.082}{\scriptsize{ (-0.023)}} \\
	
	& \textcolor[rgb]{0.002, 0.465, 0.71}{\ding{51}} & \textcolor[rgb]{0.002, 0.465, 0.71}{\ding{51}} & \textcolor[rgb]{0.002, 0.465, 0.71}{\ding{51}} & \textcolor[rgb]{0.002, 0.465, 0.71}{\ding{51}} & \textcolor[rgb]{0.847, 0.192, 0.082}{\ding{55}} & \textcolor[rgb]{0.002, 0.465, 0.71}{\ding{51}} & 0.662\textcolor[rgb]{0.847, 0.192, 0.082}{\scriptsize{ (-0.015)}} & 0.615\textcolor[rgb]{0.002, 0.465, 0.71}{\scriptsize{ (+0.010)}} & 0.358\textcolor[rgb]{0.847, 0.192, 0.082}{\scriptsize{ (-0.006)}} \\
	
	& \textcolor[rgb]{0.002, 0.465, 0.71}{\ding{51}} & \textcolor[rgb]{0.002, 0.465, 0.71}{\ding{51}} & \textcolor[rgb]{0.002, 0.465, 0.71}{\ding{51}} & \textcolor[rgb]{0.002, 0.465, 0.71}{\ding{51}} & \textcolor[rgb]{0.002, 0.465, 0.71}{\ding{51}} & \textcolor[rgb]{0.847, 0.192, 0.082}{\ding{55}} & 0.654\textcolor[rgb]{0.847, 0.192, 0.082}{\scriptsize{ (-0.023)}} &	0.635\textcolor[rgb]{0.002, 0.465, 0.71}{\scriptsize{ (+0.030)}} & 0.346\textcolor[rgb]{0.847, 0.192, 0.082}{\scriptsize{ (-0.018)}} \\
	
	& \textcolor[rgb]{0.847, 0.192, 0.082}{\ding{55}} & \textcolor[rgb]{0.847, 0.192, 0.082}{\ding{55}} & \textcolor[rgb]{0.847, 0.192, 0.082}{\ding{55}} & \textcolor[rgb]{0.847, 0.192, 0.082}{\ding{55}} & \textcolor[rgb]{0.847, 0.192, 0.082}{\ding{55}} & \textcolor[rgb]{0.847, 0.192, 0.082}{\ding{55}} & \cellcolor{red!7}0.603\textcolor[rgb]{0.847, 0.192, 0.082}{\scriptsize{ (-0.074)}} & \cellcolor{red!7}0.730\textcolor[rgb]{0.002, 0.465, 0.71}{\scriptsize{ (+0.125)}} & \cellcolor{red!7}0.301\textcolor[rgb]{0.847, 0.192, 0.082}{\scriptsize{ (-0.063)}} \\
	
	\midrule
	
	\multirow{8}{*}{\textbf{BMCMat}} & \textcolor[rgb]{0.002, 0.465, 0.71}{\ding{51}} & \textcolor[rgb]{0.002, 0.465, 0.71}{\ding{51}} & \textcolor[rgb]{0.002, 0.465, 0.71}{\ding{51}} & \textcolor[rgb]{0.002, 0.465, 0.71}{\ding{51}} & \textcolor[rgb]{0.002, 0.465, 0.71}{\ding{51}} & \textcolor[rgb]{0.002, 0.465, 0.71}{\ding{51}} & \cellcolor[rgb]{0.885, 0.96, 1.0}\textbf{0.651}\textcolor[rgb]{0.4, 0.4, 0.4}{\scriptsize{ (Base)}} & \cellcolor[rgb]{0.885, 0.96, 1.0}\textbf{4.660}\textcolor[rgb]{0.4, 0.4, 0.4}{\scriptsize{ (Base)}} & \cellcolor[rgb]{0.885, 0.96, 1.0}\textbf{0.858}\textcolor[rgb]{0.4, 0.4, 0.4}{\scriptsize{ (Base)}}  \\
	
	& \textcolor[rgb]{0.847, 0.192, 0.082}{\ding{55}} & \textcolor[rgb]{0.002, 0.465, 0.71}{\ding{51}} & \textcolor[rgb]{0.002, 0.465, 0.71}{\ding{51}} & \textcolor[rgb]{0.002, 0.465, 0.71}{\ding{51}} & \textcolor[rgb]{0.002, 0.465, 0.71}{\ding{51}} & \textcolor[rgb]{0.002, 0.465, 0.71}{\ding{51}} & 0.645\textcolor[rgb]{0.847, 0.192, 0.082}{\scriptsize{ (-0.006)}} &	 4.751\textcolor[rgb]{0.002, 0.465, 0.71}{\scriptsize{ (+0.091)}} & 0.842\textcolor[rgb]{0.847, 0.192, 0.082}{\scriptsize{ (-0.016)}} \\
	
	& \textcolor[rgb]{0.002, 0.465, 0.71}{\ding{51}} & \textcolor[rgb]{0.847, 0.192, 0.082}{\ding{55}} & \textcolor[rgb]{0.002, 0.465, 0.71}{\ding{51}} & \textcolor[rgb]{0.002, 0.465, 0.71}{\ding{51}} & \textcolor[rgb]{0.002, 0.465, 0.71}{\ding{51}} & \textcolor[rgb]{0.002, 0.465, 0.71}{\ding{51}} & 0.642\textcolor[rgb]{0.847, 0.192, 0.082}{\scriptsize{ (-0.009)}} & 4.806\textcolor[rgb]{0.002, 0.465, 0.71}{\scriptsize{ (+0.146)}} & 0.832\textcolor[rgb]{0.847, 0.192, 0.082}{\scriptsize{ (-0.026)}} \\
	
	& \textcolor[rgb]{0.002, 0.465, 0.71}{\ding{51}} & \textcolor[rgb]{0.002, 0.465, 0.71}{\ding{51}} & \textcolor[rgb]{0.847, 0.192, 0.082}{\ding{55}} & \textcolor[rgb]{0.002, 0.465, 0.71}{\ding{51}} & \textcolor[rgb]{0.002, 0.465, 0.71}{\ding{51}} & \textcolor[rgb]{0.002, 0.465, 0.71}{\ding{51}} & 0.643\textcolor[rgb]{0.847, 0.192, 0.082}{\scriptsize{ (-0.008)}} & 4.788\textcolor[rgb]{0.002, 0.465, 0.71}{\scriptsize{ (+0.128)}} & 0.835\textcolor[rgb]{0.847, 0.192, 0.082}{\scriptsize{ (-0.023)}} \\
	
	& \textcolor[rgb]{0.002, 0.465, 0.71}{\ding{51}} & \textcolor[rgb]{0.002, 0.465, 0.71}{\ding{51}} & \textcolor[rgb]{0.002, 0.465, 0.71}{\ding{51}} & \textcolor[rgb]{0.847, 0.192, 0.082}{\ding{55}} & \textcolor[rgb]{0.002, 0.465, 0.71}{\ding{51}} & \textcolor[rgb]{0.002, 0.465, 0.71}{\ding{51}} & 0.645\textcolor[rgb]{0.847, 0.192, 0.082}{\scriptsize{ (-0.006)}} & 4.752\textcolor[rgb]{0.002, 0.465, 0.71}{\scriptsize{ (+0.092)}} & 0.842\textcolor[rgb]{0.847, 0.192, 0.082}{\scriptsize{ (-0.016)}} \\
	
	& \textcolor[rgb]{0.002, 0.465, 0.71}{\ding{51}} & \textcolor[rgb]{0.002, 0.465, 0.71}{\ding{51}} & \textcolor[rgb]{0.002, 0.465, 0.71}{\ding{51}} & \textcolor[rgb]{0.002, 0.465, 0.71}{\ding{51}} & \textcolor[rgb]{0.847, 0.192, 0.082}{\ding{55}} & \textcolor[rgb]{0.002, 0.465, 0.71}{\ding{51}} & 0.644\textcolor[rgb]{0.847, 0.192, 0.082}{\scriptsize{ (-0.007)}} &	 4.764\textcolor[rgb]{0.002, 0.465, 0.71}{\scriptsize{ (+0.104)}} & 0.840\textcolor[rgb]{0.847, 0.192, 0.082}{\scriptsize{ (-0.018)}} \\
	
	& \textcolor[rgb]{0.002, 0.465, 0.71}{\ding{51}} & \textcolor[rgb]{0.002, 0.465, 0.71}{\ding{51}} & \textcolor[rgb]{0.002, 0.465, 0.71}{\ding{51}} & \textcolor[rgb]{0.002, 0.465, 0.71}{\ding{51}} & \textcolor[rgb]{0.002, 0.465, 0.71}{\ding{51}} & \textcolor[rgb]{0.847, 0.192, 0.082}{\ding{55}} & 0.638\textcolor[rgb]{0.847, 0.192, 0.082}{\scriptsize{ (-0.013)}} & 4.852\textcolor[rgb]{0.002, 0.465, 0.71}{\scriptsize{ (+0.192)}} & 0.824\textcolor[rgb]{0.847, 0.192, 0.082}{\scriptsize{ (-0.034)}} \\
	
	& \textcolor[rgb]{0.847, 0.192, 0.082}{\ding{55}} & \textcolor[rgb]{0.847, 0.192, 0.082}{\ding{55}} & \textcolor[rgb]{0.847, 0.192, 0.082}{\ding{55}} & \textcolor[rgb]{0.847, 0.192, 0.082}{\ding{55}} & \textcolor[rgb]{0.847, 0.192, 0.082}{\ding{55}} & \textcolor[rgb]{0.847, 0.192, 0.082}{\ding{55}} & \cellcolor{red!7}0.617\textcolor[rgb]{0.847, 0.192, 0.082}{\scriptsize{ (-0.034)}} & \cellcolor{red!7}5.302\textcolor[rgb]{0.002, 0.465, 0.71}{\scriptsize{ (+0.642)}} & \cellcolor{red!7}0.754\textcolor[rgb]{0.847, 0.192, 0.082}{\scriptsize{ (-0.104)}} \\
	
	\bottomrule
	
	\end{tabular}
	\vspace*{-4.4mm}
\end{table*}

Figs. \ref{eff: acc} and \ref{eff: top-acc} present the selection accuracy and top-$n$ accuracy, where RAF achieves SOTA performance on both SuiteSparse and BMCMat.
For selection accuracy, RAF outperforms existing models by margins of 0.039-0.048 on SuiteSparse and 0.022-0.055 on BMCMat with advanced feature extraction and learning mechanisms, directly translating to reduced solution times in practical applications.
As further illustrated in Fig. \ref{eff: top-acc}, RAF consistently maintains its advantage in top-$n$ selection accuracy, indicating superior selection quality and enhanced reliability for method selection.

\subsection{Ablation Study}

Compared to conventional image-based method selection, the key innovation of RAF lies in the extraction and fusion of relative image representations with six absolute numerical values, enabling more precise matrix characterization and improved selection accuracy.
We conducted an ablation study to quantify the contribution of each numerical value to RAF performance.
Experimental results evaluated on SuiteSparse and BMCMat are presented in Table \ref{ab: table} and Fig. \ref{ab: fig}.
Overall, removing any single numerical value degraded model performance, albeit to varying degrees.
Compared to the complete RAF, removing a single numerical value decreased selection accuracy (top-1) by 0.006-0.030, increased solution time by 0.01s-0.19s, reduced computational efficiency by 0.98\%-7.07\%, and decreased slowdown by 0.004-0.034 across both datasets.
These observations confirm that each numerical value in RAF contributes positively to model performance, collectively forming a complementary feature representation system.
Furthermore, removing all numerical values significantly degraded model performance, resulting in performance below that of a conventional CNN.
This underperformance occurs because, without numerical values, RAF is reduced to a conventional CNN lacking the blue channel features, further validating the contribution of matrix dimension ($N_\mathcal{A}$) features to method selection.

\section{Related Work}

Early method selection predominantly utilized machine learning.
Initially, Alternating Decision Trees were applied for method selection leveraging matrix features \cite{bhowmick2006application}.
A subsequent study employed Decision Trees and proposed three strategies based on composite methods \cite{eijkhout2010machine}.
For transient simulations, WEKA and MULAN were used, implementing K-Nearest Neighbors, Random Forests, and Decision Trees \cite{eller2012dynamic}.
The Lighthouse framework integrated several machine learning algorithms, including BayesNet and Random Forest, to select methods from the PETSc and Trilinos libraries \cite{motter2015lighthouse}.

Advances in deep learning have spurred research into its application for method selection, surpassing traditional machine learning due to its superior feature learning and handling of non-linear relationships.
Early studies employed FC for solver selection with 18 matrix features \cite{funk2022prediction}.
Subsequent research modeled matrices as topological graphs with five node and ten graph features, using GNN for method selection \cite{tang2022graph}.
Current SOTA approaches encode matrices as images, employing CNN to capture spatial patterns for method selection \cite{yamada2018preconditioner, souza2023comparison}.
In contrast, RAF enhances image-based approaches by extracting and fusing relative image representations with absolute numerical values, thereby comprehensively characterizing matrices and unlocking the potential of image-based iterative method selection.

\section{Conclusion}

In this paper, we introduce RAF, an efficient feature extraction technique that enhances image-based iterative method selection for solving sparse linear systems.
RAF simultaneously extracts image representations as relative features and corresponding numerical values as absolute features, fusing them to enhance matrix representations, thereby improving selection accuracy and accelerating linear system solutions.
Additionally, we developed BMCMat, a balanced matrix dataset constructed through PDE discretization to facilitate method selection research.
We comprehensively evaluated RAF on both SuiteSparse and BMCMat, demonstrating its SOTA performance, with improved selection accuracy by 0.02-0.06 and reduced solution times by 0.08s-0.79s, yielding 1.06x-1.17x higher computational efficiency compared to existing method selection approaches.

\bibliographystyle{IEEEtran}
\bibliography{reference}

\end{document}